\documentclass[runningheads]{llncs}

\usepackage{amssymb}
\usepackage{graphicx}
\usepackage{subfig}
\usepackage[cmex10]{amsmath}
\usepackage{lineno}
\usepackage{array}
\renewcommand{\vec}[1]{\mathbf{#1}}
\usepackage{multirow}

\usepackage{url}
\urldef{\mailsa}\path|{sxd7257, calbme}@rit.edu|

\def\registered{{\ooalign{\hfil\raise .00ex\hbox{\scriptsize R}\hfil\crcr\mathhexbox20D}}}

\begin{document}

\mainmatter  


\title{Left Ventricle Segmentation and Quantification from Cardiac Cine MR Images via Multi-task Learning}

\titlerunning{\it LV Segmentation and Quantification via Multi-task Learning}

%
%
%
\author{Shusil Dangi${^1}$ \and Ziv Yaniv${^{2,3}}$ \and Cristian A. Linte${^{1,4}}$}

\authorrunning{Dangi, S. {\it et al.}}

\institute{${^1}$Center for Imaging Science, Rochester Institute of Technology, Rochester NY USA \\ 
${^2}$TAJ Technologies Inc., Bloomington MN USA \\
${^3}$National Library of Medicine, National Institutes of Health, Bethesda MD USA \\
${^4}$Biomedical Engineering, Rochester Institute of Technology, Rochester NY USA \\
\mailsa
}


\maketitle
%
\begin{abstract}
Segmentation of the left ventricle and quantification of various cardiac contractile functions is crucial for the timely diagnosis and treatment of cardiovascular diseases. Traditionally, the two tasks have been tackled independently. Here we propose a convolutional neural network based multi-task learning approach to perform both tasks simultaneously, such that, the network learns better representation of the data with improved generalization performance. Probabilistic formulation of the problem enables learning the task uncertainties during the training, which are used to automatically compute the weights for the tasks. We performed a five fold cross-validation of the myocardium segmentation obtained from the proposed multi-task network on 97 patient 4-dimensional cardiac cine-MRI datasets available through the STACOM LV segmentation challenge against the provided gold-standard myocardium segmentation, obtaining a Dice overlap of $0.849 \pm 0.036$ and mean surface distance of $0.274 \pm 0.083$ mm, while simultaneously estimating the myocardial area with mean absolute difference error of $205\pm198$ mm$^2$.
\end{abstract}

\section{Introduction}
\label{sec:Intro}
Magnetic Resonance Imaging (MRI) is the preferred imaging modality for non-invasive assessment of cardiac performance, thanks to its lack of ionizing radiation, good soft tissue contrast, and high image quality. Cardiac contractile function parameters such as systolic and diastolic volumes, ejection fraction, and myocardium mass are good indicators of cardiac health, representing reliable diagnostic value. Segmentation of the left ventricle (LV) allows us to compute these cardiac parameters, and also to generate high quality anatomical models for surgical planning, guidance, and regional analysis of the heart. Although manual delineation of the ventricle is considered as the gold-standard, it is time consuming and highly susceptible to inter- and intra-observer variability. Hence, there is a need for fast, robust, and accurate semi- or fully-automatic segmentation algorithms.

Cardiac MR image segmentation techniques can be broadly classified into: 1) no-prior based methods, such as thresholding, edge-detection and linking, and region growing; 2) weak-prior based methods, such as active contours (snakes), level-set, and graph-theoretical models; 3) strong-prior based methods, such as active shape and appearance models, and atlas-based models; and 4) machine learning based methods, such as per pixel classification and convolutional neural network (CNN) based models. A comprehensive review of various cardiac MR image segmentation techniques can be found in \cite{Petitjean2011}.

Recent success of deep learning techniques \cite{LeCun:2015} in high level computer vision, speech recognition, and natural language processing applications has motivated their use in medical image analysis. Although the early adoption of deep learning in medical image analysis encountered various challenges, such as the limited availability of medical imaging data and associated costly manual annotation, those challenges were circumvented by patch-based training, data augmentation, and transfer learning techniques \cite{Shen:2017,Litjens:2017}.

Long {\it et al.} \cite{Long:2015} were the first to propose a fully convolutional network (FCN) for semantic image segmentation by adapting the contemporary classification networks fine-tuned for the segmentation task, obtaining state-of-the-art results. Several modifications to the FCN architecture and various post-processing schemes have been proposed to improve the semantic segmentation results as summarized in \cite{Garcia:2017}. Notably, the U-Net architecture \cite{Ronneberger:2015} with data augmentation has been very successful in medical image segmentation. 

While segmentation indirectly enables the computation of various cardiac indices, direct estimation of these quantities from low-dimensional representation of the image have also been proposed in the literature \cite{Xiantong:2015,Xue:2017,Xue:2018}. However, these methods are less interpretable and the correctness of the produced output is often unverifiable, potentially limiting their clinical adoption. 

Here we propose a CNN based multi-task learning approach to perform both LV segmentation and cardiac indices estimation simultaneously, such that these related tasks regularize the network, hence improving the network generalization performance. Furthermore, our method increases the interpretablity of the output cardiac indices, as the clinicians can infer its correctness based on the quality of produced segmentation result.


\section{Methodology}
\label{sec:Methods}
Traditionally, the segmentation of the LV and quantification of the cardiac indices have been performed independently. However, due to a close relation between the two tasks, we identified that learning a CNN model to perform both tasks simultaneously is beneficial in two ways: 1) it forces the network to learn features important for both tasks, hence, reducing the chances of over-fitting to a specific task, improving generalization; 2) the segmentation results can be used as a proxy to identify the reliability of the obtained cardiac indices, and also to perform regional cardiac analysis and surgical planning. 

\subsection{Data Preprocessing and Augmentation}
\label{ssec:DataPreprocessing}
This study employed 97 de-identified cardiac MRI image datasets from patients suffering from myocardial infarction and impaired LV contraction available as a part of the STACOM Cardiac Atlas Segmentation Challenge project \cite{Fonseca:2011,Suinesiaputra:2014} database\footnote{\url{http://www.cardiacatlas.org}}. Cine-MRI images in short-axis and long-axis views are available for each case. The semi-automated myocardium segmentation provided with the dataset served as gold-standard for assessing the proposed segmentation technique. The dataset was divided into 80\% training and 20\% testing for five-fold cross-validation.

The physical pixel spacing in the short-axis images ranged from 0.7031 to 2.0833 mm. We used SimpleITK~\cite{Yaniv:2017} to resample all images to the most common spacing of $1.5625$ mm along both x- and y-axis. The resampled images were center cropped or zero padded to a common resolution of $192\times192$ pixels. We applied two transformations, obtained from the combination of random rotation and translation (by maximum of half the image size along x- and y-axis), to each training image for data augmentation.

\subsection{Multi-Task Learning using Uncertainty to Weigh Losses}
We estimate the task-dependent uncertainty \cite{Kendall:2017} for both myocardium segmentation and myocardium area regression via probabilistic modeling. The weights for each task are determined automatically based on the task uncertainties learned during the training \cite{Kendall:2017MTL}.

For a neural network with weights $\vec{W}$, let $\vec{f}^{\vec{W}}(\vec{x})$ be the output for the corresponding input $\vec{x}$. We model the likelihood for segmentation task as the squashed and scaled version of the model output through a softmax function:
\begin{equation}
p(\vec{y}|\vec{f}^{\vec{W}}(\vec{x}),\sigma) = \text{Softmax}\left(\frac{1}{\sigma^2}\vec{f}^{\vec{W}}(\vec{x})\right)
\end{equation}
where, $\sigma$ is a positive scalar, equivalent to the {\it temperature}, for the defined Gibbs/Boltzmann distribution. The magnitude of $\sigma$ determines the {\it uniformity} of the discrete distribution, and hence relates to the uncertainty of the prediction. The log-likelihood for the segmentation task can be written as:
\begin{equation}
\text{log}~p(\vec{y}=c|\vec{f}^{\vec{W}}(\vec{x}),\sigma) = \frac{1}{\sigma^2}f_{c}^{\vec{W}}(\vec{x})-\text{log}\sum_{c'}\text{exp}\left(\frac{1}{\sigma^2}f_{c'}^{\vec{W}}(\vec{x})\right)
\end{equation}
where $f_{c}^{\vec{W}}(\vec{x})$ is the $c$'th element of the vector $\vec{f}^{\vec{W}}(\vec{x})$.

Similarly, for the regression task we define our likelihood as a Lapacian distribution with its mean given by the neural network output:
\begin{equation}
p(\vec{y}|\vec{f}^{\vec{W}}(\vec{x}),\sigma) = \frac{1}{2\sigma}\text{exp}\left(-\frac{|\vec{y}-\vec{f}^{\vec{W}}(\vec{x})|}{\sigma}\right)
\end{equation}
The log-likelihood for regression task can be written as:
\begin{equation}
\text{log}~p(\vec{y}|\vec{f}^{\vec{W}}(\vec{x}),\sigma) \propto -\frac{1}{\sigma}|\vec{y}-\vec{f}^{\vec{W}}(\vec{x})|-\text{log}\sigma
\end{equation}
where $\sigma$ is the neural networks observation noise parameter --- capturing the noise in the output.

For a network with two outputs: continuous output $\vec{y}_1$ modeled with a Laplacian likelihood, and a discrete output $\vec{y}_2$ modeled with a softmax likelihood, the joint loss $\mathcal{L}(\vec{W},\sigma_1,\sigma_2)$ is given by:
\begin{equation}
\begin{split}
\mathcal{L}(\vec{W},\sigma_1,\sigma_2) & = -\text{log}~p(\vec{y}_1,\vec{y}_2=c|\vec{f}^{\vec{W}}(\vec{x})) \\
& = -\text{log}~\left(p(\vec{y}_1|\vec{f}^{\vec{W}}(\vec{x}),\sigma_1) \cdot p(\vec{y}_2=c|\vec{f}^{\vec{W}}(\vec{x}),\sigma_2)\right) \\
& \approx \frac{1}{\sigma_1} \mathcal{L}_1(\vec{W}) + \frac{1}{\sigma_2^2} \mathcal{L}_2(\vec{W})+ \text{log} \sigma_1 + \text{log} \sigma_2
\end{split}
\label{eq:jointloss}
\end{equation}
where $\mathcal{L}_1(\vec{W})=|\vec{y_{1}}-\vec{f}^{\vec{W}}(\vec{x})|$ is the mean absolute distance (MAD) loss of $\vec{y}_1$ and $\mathcal{L}_2(\vec{W})=-\text{log}~\text{Softmax}(\vec{y}_2,\vec{f}^{\vec{W}}(\vec{x}))$ is the cross-entropy loss of $\vec{y}_2$. To arrive at {\bf(\ref{eq:jointloss})}, the two tasks are assumed independent and simplifying assumptions have been made for the softmax likelihood, resulting in a simple optimization objective with improved empirical results \cite{Kendall:2017MTL}. During the training, the joint likelihood loss $\mathcal{L}(\vec{W},\sigma_1,\sigma_2)$ is optimized with respect to $\vec{W}$ as well as $\sigma_1$, $\sigma_2$.

As observed in {\bf(\ref{eq:jointloss})}, the uncertainties ($\sigma_1$, $\sigma_2$) learned during the training are weighting the losses for individual tasks, such that, the task with higher uncertainty is weighted less and vice versa. Furthermore, the uncertainties can't become too large, as they are penalized by the last two terms in {\bf(\ref{eq:jointloss})}.

\begin{figure}[!h]
\centering
\includegraphics[width = \columnwidth]{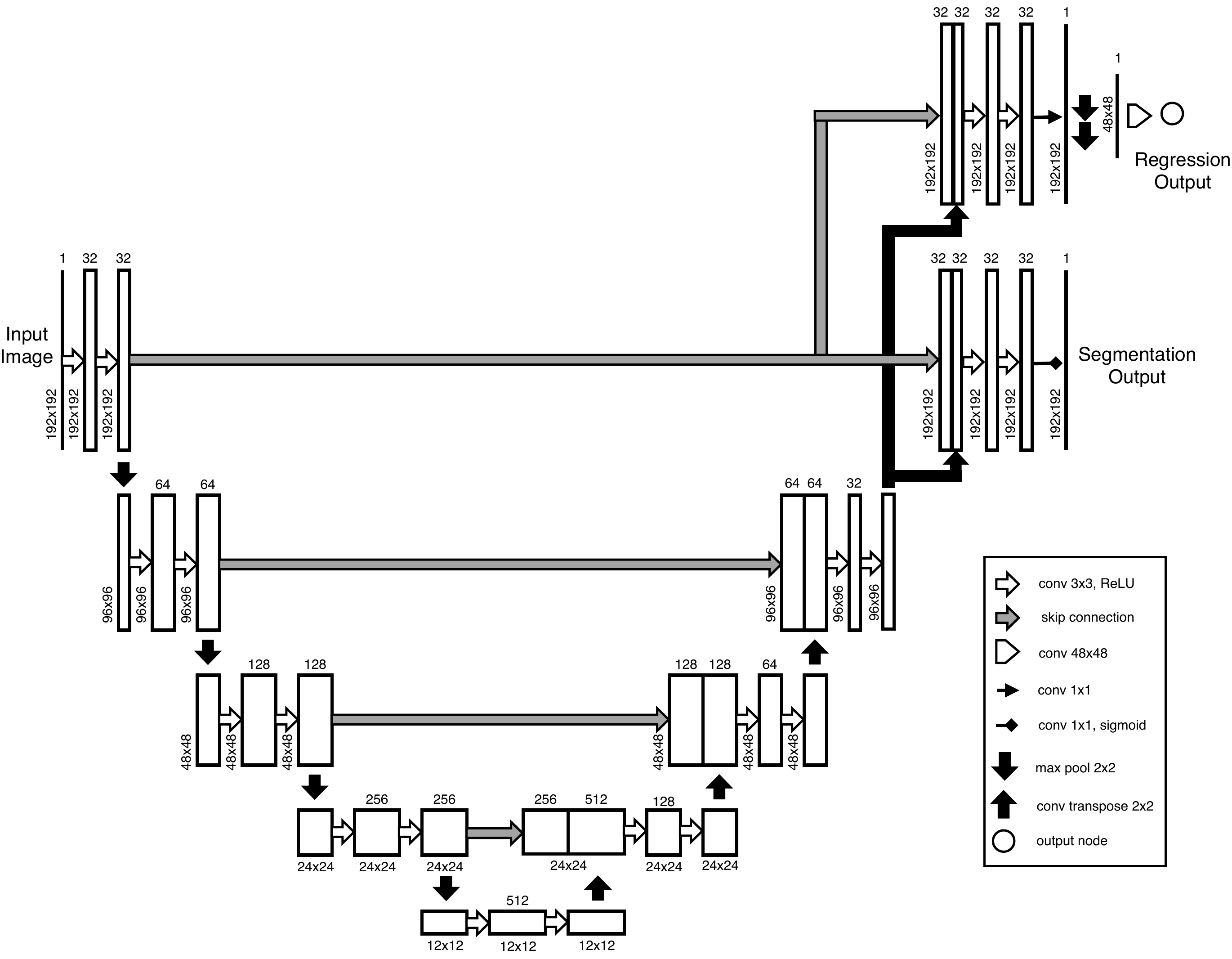}
\caption{\scriptsize{Modified U-Net architecture for multi-task learning. The segmentation and regression tasks are split at the final up-sampling and concatenation layer. The final feature map in the segmentation path is passed through a sigmoid layer to obtain a per-pixel image segmentation. Similarly, the final feature map in the regression path is down-sampled (by max-pooling) to $1/4^{th}$ of its size and fed to a fully-connected layer to generate a single regression output. The logarithm of the task uncertainties ($\text{log}\sigma_1$, $\text{log}\sigma_2$) are set as network parameters and are encoded in the loss function {\bf(\ref{eq:jointloss})}, hence learned during the training.}}
\label{fig:NetworkArchitecture}
\end{figure}

\subsection{Network Architecture}
In this work, we adapt the U-Net architecture \cite{Ronneberger:2015}, highly successful in medical image segmentation, to perform an additional task of myocardium area estimation as shown in {\bf Fig. \ref{fig:NetworkArchitecture}}. The segmentation and regression paths are split at the final up-sampling and concatenation layer. The final feature map in the segmentation path is passed through a sigmoid layer to obtain a per-pixel image segmentation. Similarly, the regression output is obtained by down-sampling the final feature map in the regression path by $1/4^{th}$ of its size and passing it through a fully-connected layer. The logarithm of the task uncertainties ($\text{log}\sigma_1$, $\text{log}\sigma_2$) added as the network parameters are used to construct the loss function {\bf(\ref{eq:jointloss})}, and are learned during the training. Note that we train the network to predict the log uncertainty $s = \text{log} (\sigma$) due to its numerical stability and the positivity constraint imposed on the computed uncertainty via exponentiation, $\sigma = \text{exp}(s)$.

\section{Results}
The network was initialized with the {\it Kaiming uniform} \cite{He:2015} initializer and trained for 50 epochs using {\it RMS prop} optimizer with a learning rate of 0.001 (decayed by 0.95 every epoch) in PyTorch\footnote{https://github.com/pytorch/pytorch}. The best performing network, in terms of the Dice overlap between the obtained and gold-standard segmentation, in the test set, was saved and used for evaluation. 

The network training required 9 minutes per epoch on average using a 12GB Nvidia Titan Xp GPU. It takes 0.663 milliseconds on average to process a slice during testing. The log uncertainties learned for the segmentation and regression tasks during training are $-3.9$ and $3.45$, respectively, which correspond to weighting the cross-entropy and mean absolute difference (MAD) loss by a ratio of 1556:1. Note that the scale for cross-entropy loss is $10^{-2}$, whereas that for MAD loss is $10^2$.

\scriptsize
\begin{table}[!h]
{\caption{\scriptsize{Evaluation of the segmentation results obtained from the baseline U-Net (UNet) architecture and the proposed multi-task network (MTN) against the provided gold-standard myocardium segmentation using --- Dice Index, Jaccard Index, Mean Surface Distance, and Hausdorff Distance.}}
\label{tab:Segmentation}}
\begin{center}
\begin{tabular}{|p{2.5cm}||>{\centering\arraybackslash}m{1.4cm}|>{\centering\arraybackslash}m{1.4cm}||>{\centering\arraybackslash}m{1.4cm}|>{\centering\arraybackslash}m{1.4cm}||>{\centering\arraybackslash}m{1.4cm}|>{\centering\arraybackslash}m{1.4cm}|}
\hline
\multirow{2}{2.5cm}{\bf Assessment Metric} & \multicolumn{2}{|c||}{\bf End-Diastole} & \multicolumn{2}{|c||}{\bf End-Systole} & \multicolumn{2}{|c|}{\bf All Phases}\\ \cline{2-7}
& UNet & {\bf MTN} & UNet & {\bf MTN} & UNet & {\bf MTN} \\
\hline \hline
{\bf Dice Index} & $0.836\pm0.036$ & ${\bf 0.837}\pm0.038$ & ${\bf 0.850}\pm0.033$ & $0.849\pm0.036$ & $0.847\pm0.035$ & ${\bf 0.849}\pm0.036$ \\
\hline
{\bf Jaccard Index} & $0.719\pm0.052$ & ${\bf 0.721}\pm0.054$ & ${\bf 0.740}\pm0.048$ & $0.739\pm0.053$ & $0.736\pm0.050$ & ${\bf 0.739}\pm0.053$ \\
\hline
{\bf Mean Surface Distance (mm)} & $0.318\pm0.089$ & ${\bf 0.286}\pm0.087$ & $0.299\pm0.095$ & ${\bf 0.274}\pm0.090$ & $0.305\pm0.088$ & ${\bf 0.274}\pm0.083$ \\
\hline
{\bf Hausdorff Distance (mm)} & $13.582\pm4.337$ & ${\bf 13.364}\pm4.108$ & ${\bf 13.083}\pm3.630$ & $13.355\pm3.861$ & ${\bf 13.211}\pm4.212$ & $13.233\pm3.810$ \\
\hline
\end{tabular}
\end{center}
\end{table}
\normalsize

\begin{figure}[!h]
\centering
\includegraphics[width = \columnwidth]{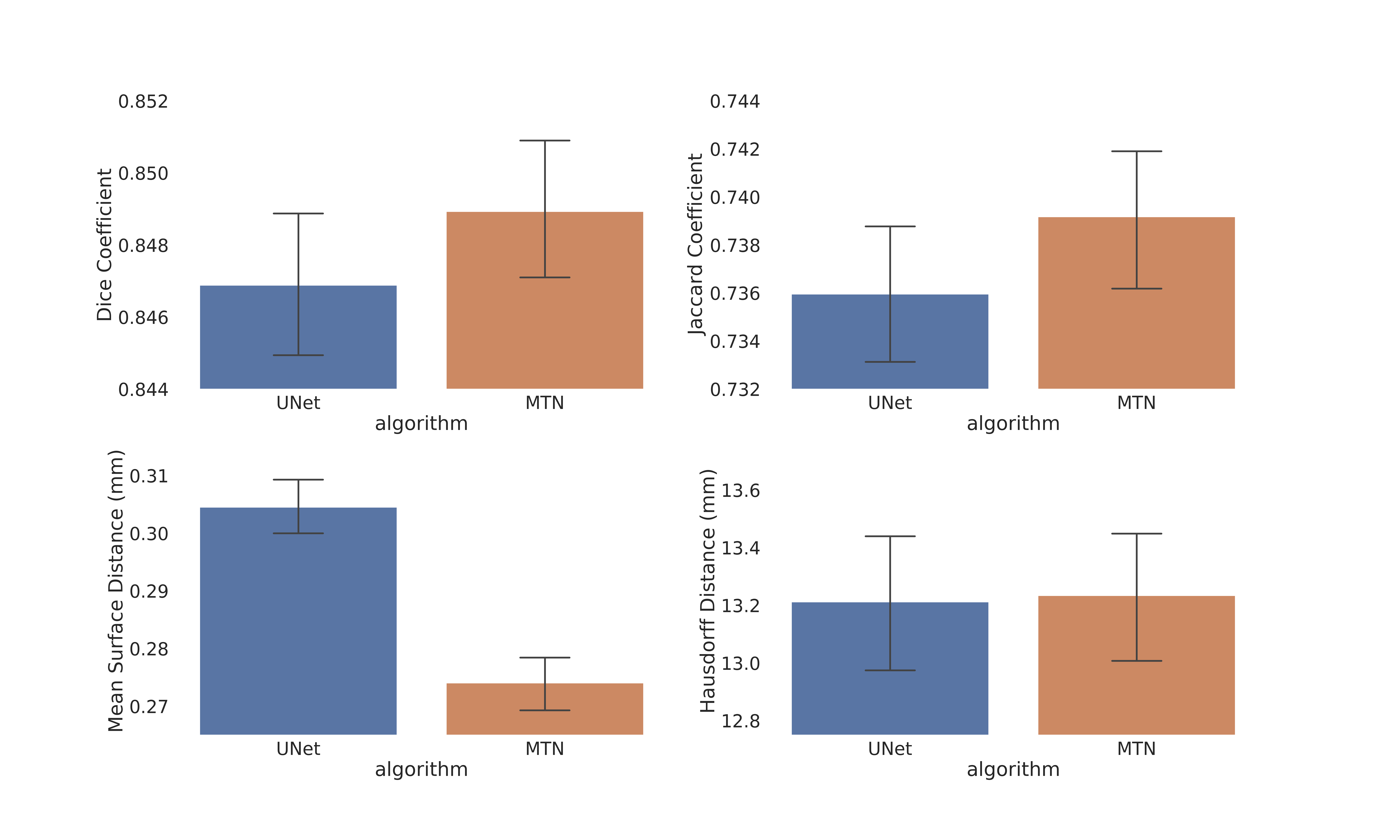}
\caption{\scriptsize{Mean and 99\% confidence interval for (a) Dice Coefficient, (b) Jaccard Coefficient, (c) Mean Surface Distance (mm), and (d) Hausdorff Distance (mm), for baseline U-Net and the proposed MTN architecture across all cardiac phases. Confidence interval is obtained based on 1000 bootstrap re-sampling with replacement for 2191 test volumes across five-fold cross-validation.}}
\label{fig:SegmentationComparisonCI}
\end{figure}

The 2D segmentation results are stacked to form a 3D volume, and the largest connected component is selected as the final myocardium segmentation. The myocardium segmentation obtained for end-diastole, end-systole, and all cardiac phases from the proposed multi-task network (MTN) and from the baseline U-Net architecture (without the regression task) are both assessed against the gold-standard segmentation provided with the dataset as part of the challenge, using four traditionally employed segmentation metrics --- Dice Index, Jaccard Index, Mean surface distance (MSD), and Hausdorff distance (HD) --- summarized in {\bf Table \ref{tab:Segmentation}}. Note that the myocardium dice coefficient is higher for end-systole phase where the myocardium is thickest.

The Kolmogorov-Smirnov test shows that the difference in distributions for Dice, Jaccard and MSD metrics between the proposed multi-task network and baseline U-Net architecture are statistically significant with p-values: $2.156e^{-4}$, $2.156e^{-4}$, and $6.950e^{-34}$, respectively. 
However, since the segmentation is evaluated on a large sample of 2191 volumes across five-fold cross validation, the p-values quickly go to zero even for slight difference in distributions being compared, representing no practical significance \cite{Lin:2013}. Hence, we computed the 99\% confidence interval for the mean value of each segmentation metric based on 1000 bootstrap re-sampling with replacement, as shown in {\bf Fig. \ref{fig:SegmentationComparisonCI}}. As evident from {\bf Fig. \ref{fig:SegmentationComparisonCI}}, Dice, Jaccard and HD metrics are statistically similar, whereas the reduction in MSD for the proposed multi-task network compared to the baseline U-Net architecture is statistically significant.

\begin{figure}[!h]
\centering
\includegraphics[width = \columnwidth]{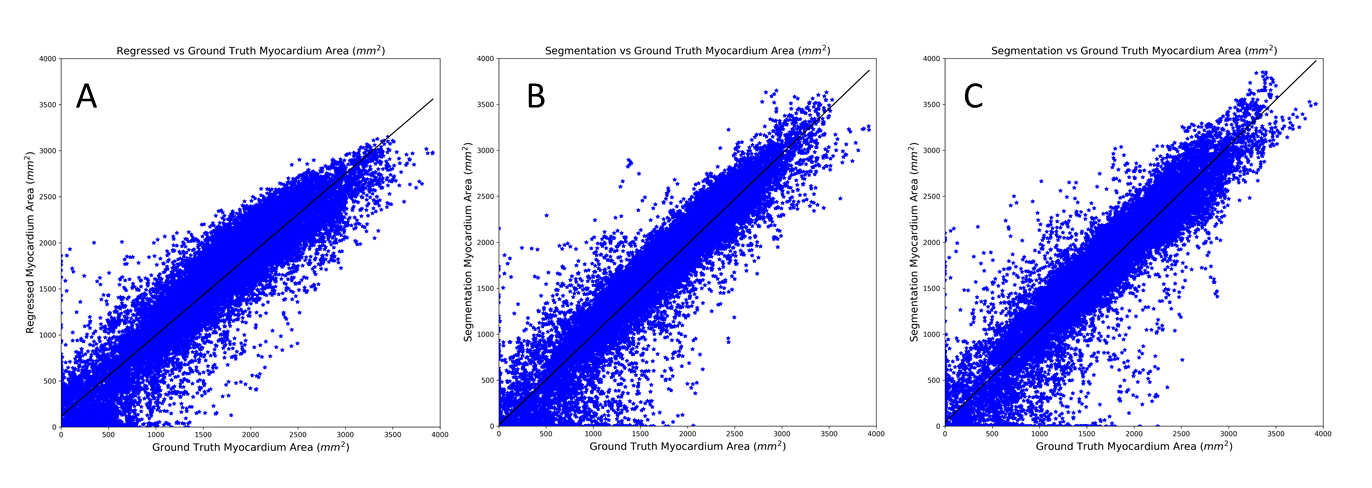}
\caption{\scriptsize{The myocardium area computed from A) regression path of the proposed multi-task network, B) segmentation obtained from the proposed multi-task network, C) segmentation obtained from the baseline U-Net model, plotted against the corresponding myocardium area obtained from the provided gold-standard segmentation for all cardiac phases. The best fit line is shown in each plot. The correlation coefficients for A, B, and C are 0.9466, 0.9565, 0.9518, respectively}}
\label{fig:RegressionPlot}
\end{figure}

In addition to obtaining the myocardium area from the regression path of the proposed network, it can also be computed indirectly from the obtained myocardium segmentation. Hence, we compute and evaluate the myocardium area estimated from three different sources: (a) regression path of the MTN, (b) segmentation obtained from the MTN, and (c) segmentation obtained from the baseline U-Net model. {\bf Fig. \ref{fig:RegressionPlot}} shows the myocardium area obtained from these three methods for all phases of the cardiac cycle plotted against the ground-truth myocardium area estimated from the gold-standard myocardium segmentation provided as part of the challenge data. We can observe a linear relationship between the computed and gold-standard myocardium areas, and the corresponding correlation coefficients for the methods (a), (b), and (c) are 0.9466, 0.9565, 0.9518, respectively.

\scriptsize
\begin{table}[!h]
{\caption{\scriptsize{Mean absolute difference (MAD), in mm$^2$, between the myocardium area obtained from the provided gold-standard segmentation and the results computed from: (a) the regression path of the proposed multi-task network, (b) segmentation obtained from the proposed multi-task network, and (c) segmentation obtained from the baseline U-Net model, for end-diastole, end-systole, and all cardiac phases, sub-divided into apical, mid, and basal regions of the heart.}}
\label{tab:MyoArea}}
\begin{center}
\begin{tabular}{|m{1.3cm}||>{\centering\arraybackslash}m{1cm}|>{\centering\arraybackslash}m{1cm}|>{\centering\arraybackslash}m{1cm}||>{\centering\arraybackslash}m{1cm}|>{\centering\arraybackslash}m{1cm}|>{\centering\arraybackslash}m{1cm}||>{\centering\arraybackslash}m{1cm}|>{\centering\arraybackslash}m{1cm}|>{\centering\arraybackslash}m{1cm}|}
\hline
\multirow{2}{1.5cm}{\bf Cardiac Regions} & \multicolumn{3}{|c||}{\bf End-Diastole} & \multicolumn{3}{|c||}{\bf End-Systole} & \multicolumn{3}{|c|}{\bf All Phases}\\ \cline{2-10}
& {\bf Reg-MTN} & {\bf Seg-MTN} & Seg-UNet & {\bf Reg-MTN} & {\bf Seg-MTN} & Seg-UNet & {\bf Reg-MTN} & {\bf Seg-MTN} & Seg-UNet \\
\hline \hline
{\bf All} & $201\pm199$ & ${\bf 174}\pm209$ & $203\pm221$ & $211\pm209$ & ${\bf 173}\pm203$ & $187\pm204$ & $206\pm198$ & ${\bf 170}\pm199$ & $193\pm208$ \\ \hline
{\bf Apical} & ${\bf 185}\pm180$ & $187\pm204$ & $194\pm186$ & $193\pm199$ & $190\pm226$ & ${\bf 185}\pm185$ & $184\pm183$ & ${\bf 181}\pm210$ & $187\pm189$ \\ \hline
{\bf Mid} & $190\pm172$ & ${\bf 141}\pm132$ & $179\pm142$ & $228\pm194$ & ${\bf 160}\pm151$ & $174\pm135$ & $212\pm178$ & ${\bf 149}\pm132$ & $176\pm141$ \\ \hline
{\bf Basal} & ${\bf 250}\pm269$ & $252\pm331$ & $282\pm368$ & $193\pm241$ & ${\bf 186}\pm267$ & $216\pm312$ & $213\pm248$ & ${\bf 210}\pm289$ & $237\pm319$ \\ \hline
\end{tabular}
\end{center}
\end{table}
\normalsize

\begin{figure}[!h]
\centering
\subfloat[\scriptsize{Box-plot for the mean absolute difference (MAD). \label{fig:MAEComparisonA}}]{
\includegraphics[width = \columnwidth]{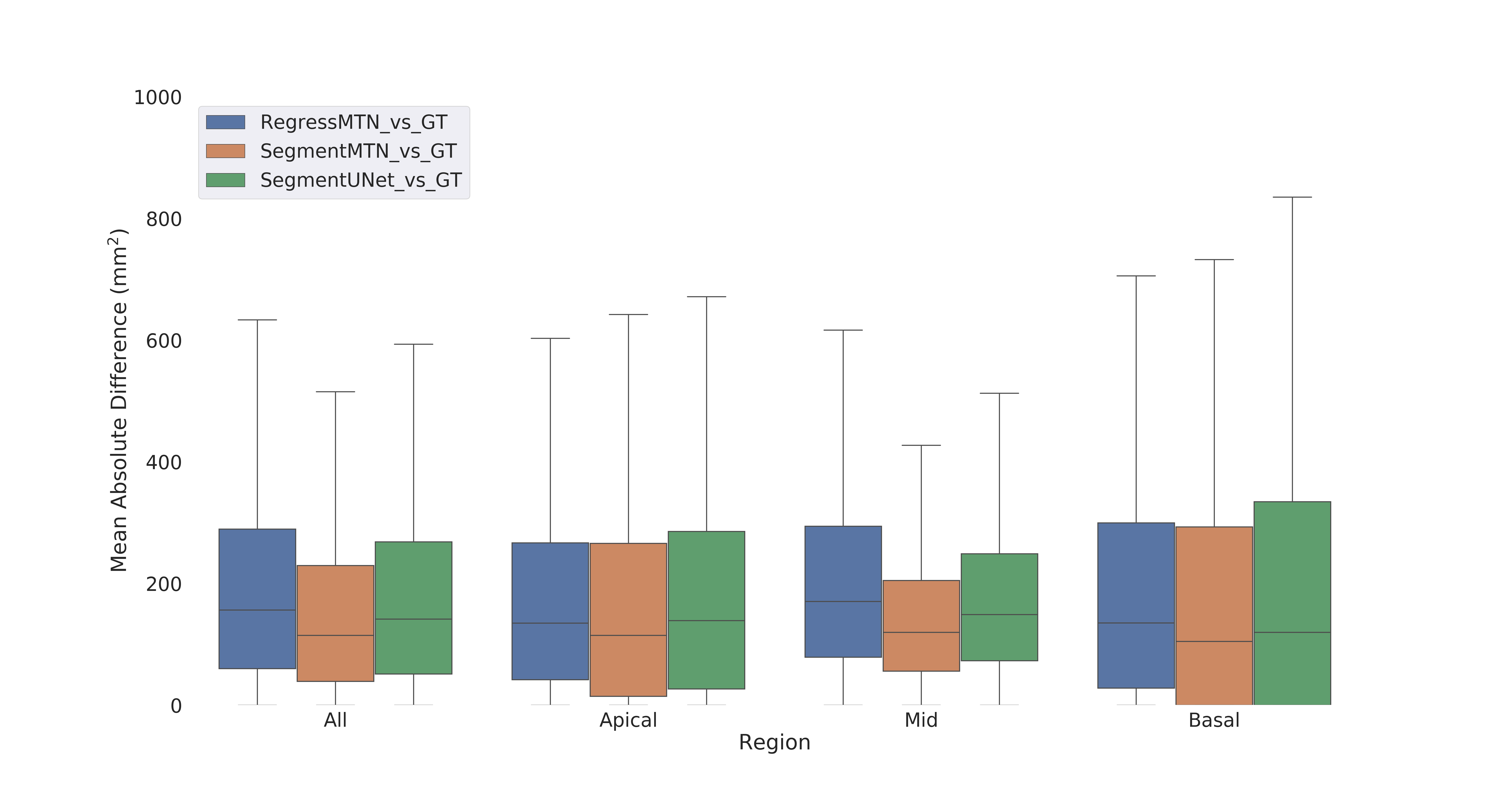}
}

\subfloat[\scriptsize{Mean and 99\% confidence interval for the mean absolute difference (MAD). \label{fig:MAEComparisonB}}]{
\includegraphics[width = \columnwidth]{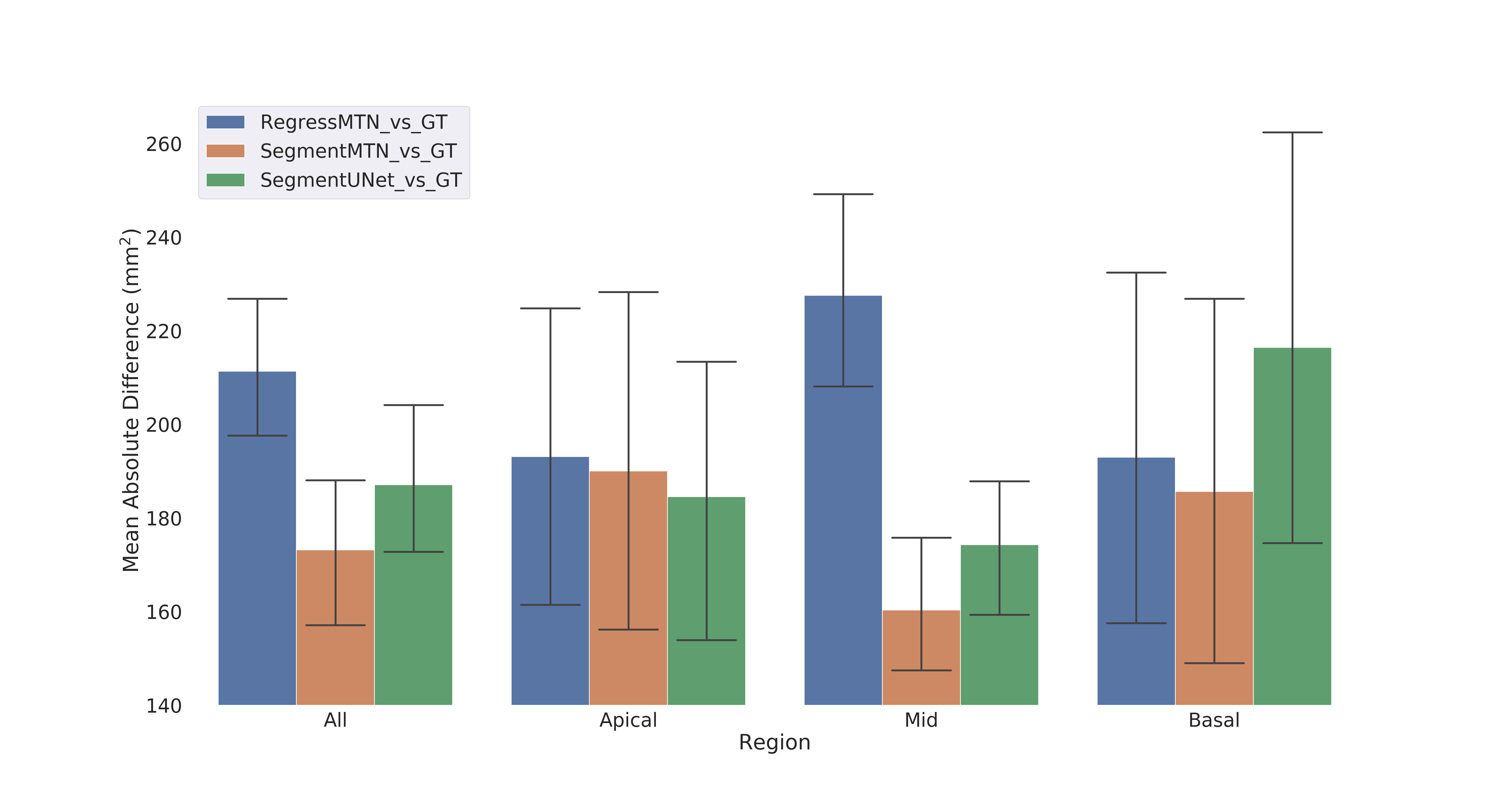}
}
\caption{\scriptsize{(a) Box-plot (outliers removed for clarity) and (b) Mean and 99\% confidence interval, for the mean absolute difference (MAD) between the myocardium area obtained from the provided gold-standard segmentation and the results obtained from: (1) the regression path of the proposed multi-task network, (2) segmentation obtained from the proposed multi-task network, and (3) segmentation obtained from the baseline U-Net model. Confidence intervals are obtained based on 1000 bootstrap re-sampling with replacement.}}
\label{fig:MAEComparison}
\end{figure}


Further, we computed the MAD between the ground-truth myocardium area and the area estimated by each of the three methods for end-diastole, end-systole, and all cardiac phases (for 26664 slices across five-fold cross validation). For the regional analysis, slices in the ground-truth segmentation after excluding two apical and two basal slices are considered as mid-slices. {\bf Table \ref{tab:MyoArea}} summarizes the mean and standard deviation for the computed MADs. Box-plots (outliers removed for clarity) comparing the three methods for different regions of the heart throughout the cardiac cycle are shown in {\bf Fig. \ref{fig:MAEComparisonA}}. The MAD in myocardium area estimation of $206 \pm 198$ mm$^2$ obtained from the regression output of the proposed method is similar to the results presented in \cite{Xue:2017}: $223\pm193$ mm$^2$, while acknowledging the limitation that the study in \cite{Xue:2017} was conducted on a different dataset than our study. Moreover, while the regression output of the proposed network yields good estimate of the myocardial area, the box-plot in {\bf Fig. \ref{fig:MAEComparisonA}} suggests that even further improved myocardial area estimates can be obtained from a segmentation based method, provided that the quality of the segmentation is good. 

Lastly, we computed the 99\% confidence interval for the mean value of myocardium area MAD based on 1000 bootstrap re-sampling with replacement, as shown in {\bf Fig. \ref{fig:MAEComparisonB}}. This confirms that the myocardium area estimated from the segmentation output of the proposed multi-task network is significantly better than that obtained from the regression output, however, there is no statistical significance between other methods. Furthermore, we can observe the variability in MAD is highest for the basal slices, followed by apical and mid slices.

\section{Discussion, Conclusion, and Future Work}
We presented a multi-task learning approach to simultaneously segment and quantify myocardial area. We adapt the U-Net architecture, highly successful in medical image segmentation, to perform an additional regression task. The best location to incorporate the regression path into the network is a hyper-parameter, tuned empirically. We found that adding the regression path in the bottleneck or intermediate decoder layers is detrimental for the segmentation performance of the network, likely due to high influence of the skip connections in the U-Net architecture.

Myocardium area estimates obtained from the regression path of the proposed network are similar to the direct estimation-based results found in the literature. However, our experiments suggest that segmentation-based myocardium area estimation is superior to that obtained from a direct estimation-based method. 
Lastly, the myocardium segmentation obtained from our method is at least as good as the segmentation obtained from the baseline U-Net model.

To test the generalization performance of the proposed multi-task network, we plan to evaluate the network performance using a lower number of training images. Similarly, we plan to extend this work to segment left ventricle myocardium, blood-pool, and right ventricle, and regress their corresponding areas using the Automated Cardiac Diagnosis Challenge (ACDC)\footnote{https://www.creatis.insa-lyon.fr/Challenge/acdc/} 2017 dataset.\\
\\
{\bf Acknowledgements}\\
Research reported in this publication was supported by the National Institute of General Medical Sciences of the National Institutes of Health under Award No. R35GM128877 and by the Office of Advanced Cyberinfrastructure of the National Science Foundation under Award No. 1808530. Ziv Yaniv's work was supported by the Intramural Research Program of the U.S. National Institutes of Health, National Library of Medicine.

\bibliographystyle{splncs}   
\bibliography{main}   

\end{document}